\documentclass{article}

\usepackage[nonatbib, final]{neurips_2021}
\usepackage{neurips_2021}
\usepackage{graphicx}
\usepackage{wrapfig}

\usepackage[utf8]{inputenc} 
\usepackage[T1]{fontenc}    
\usepackage{hyperref}       
\usepackage{url}            
\usepackage{booktabs}       
\usepackage{amsfonts}       
\usepackage{nicefrac}       
\usepackage{microtype}      
\usepackage{apacite}
\usepackage{caption}
\usepackage{array}
\newcolumntype{C}[1]{>{\centering\arraybackslash}p{#1}}

\usepackage{xcolor, soul}
\definecolor{Red}{RGB}{255,0,0}
\definecolor{Green}{RGB}{0,102,0}
\definecolor{Blue}{RGB}{10,100,200}
\definecolor{Purple}{RGB}{76,0,153}

 \title{Learning to solve complex tasks by growing knowledge culturally across generations}

\author{%
  Michael Henry Tessler$^{1,2}$, Jason Madeano$^{1}$, Pedro A. Tsividis$^{1,3}$, \\ 
  \textbf{Brin Harper}$^{1}$, \textbf{Noah D. Goodman}$^{4}$, and \textbf{Joshua B. Tenenbaum}$^{1}$\\
  $^{1}$MIT, $^{2}$DeepMind, $^{3}$Common Sense Machines, $^{4}$Stanford University\\
  Correspondence: \texttt{tessler@mit.edu} \\
}

\begin{document}

\maketitle

\begin{abstract}
Knowledge built culturally across generations allows humans to learn far more than an individual could glean from their own experience in a lifetime.  Cultural knowledge in turn rests on language: language is the richest record of what previous generations believed, valued, and practiced, and how these evolved over time.
The power and mechanisms of language as a means of cultural learning, however, are not well understood, and as a result, current AI systems do not leverage language as a means for cultural knowledge transmission.  Here, we take a first step towards reverse-engineering cultural learning through language. We developed a suite of complex  tasks in the form of minimalist-style video games, which we deployed in an iterated learning paradigm. Human participants were limited to only two attempts (two lives) to beat each game and were allowed to write a message to a future participant who read the message before playing. 
Knowledge accumulated gradually across generations, allowing later generations to advance further in the games and perform more efficient actions. Multigenerational learning followed a strikingly similar trajectory to individuals learning alone with an unlimited number of lives. Successive generations of learners were able to succeed by expressing distinct types of knowledge in natural language: the dynamics of the environment, valuable goals, dangerous risks, and strategies for success.
The video game paradigm we pioneer here is thus a rich test bed for developing AI systems capable of acquiring and transmitting cultural knowledge. 
\end{abstract}
\vspace{-0.1cm}
\section{Introduction}
\vspace{-0.1cm}
Innovation and discovery are critical to our species' ecological success in diverse and challenging environments. 
Impressive as it may be, 
cleverness alone is probably not enough to allow an isolated individual to endure a winter of sub-freezing temperatures and perpetual darkness in the Arctic Archipelago. 
To improve one's chances of survival in such hostile terrain, one should learn how to hunt seals and make skin clothing \cite{lambert2014gates}. 
This knowledge may be difficult to discover, but once discovered, it can be preserved and built upon by future generations if there is a mechanism for high-fidelity transmission \cite{boyd2011cultural}.
Together, innovation and faithful transmission produce a ``cultural ratchet'', the ability to make intellectual and technological progress while preventing reversion to more basic knowledge states \cite{tomasello1999cultural}.

Cumulative cultural evolution describes the gradual improvement of knowledge and technology across generations of individual learners who have a means of transmitting their knowledge \cite{boyd1988culture, tomasello1999cultural, mesoudi2018cumulative}.
The learning trajectories of some of the most powerful AI systems often follow a different route: knowledge and expertise is built up within a single, isolated system in a manner which is often inscrutable even to its human designers \cite{silver2016mastering,silver2018general}.
When AI systems have learned through an evolutionary process (e.g., OpenAI Five or DeepMind's AlphaStar; \citeNP{berner2019dota, vinyals2019grandmaster}), the process has been most similar to that of biological evolution, involving random perturbations in agent strategies and survival-of-the-fittest dynamics.
For AI systems to efficiently learn from one another, they will have to be able to describe their experiences, models, and policies in ways that go beyond existing knowledge distillation approaches \cite{yim2017gift}. Furthermore, for AI systems to be incorporated into the cultural evolutionary process with humans \cite{dafoe2020open}, they will have to be able to describe their learned knowledge in a way that humans can understand and build on.
AI systems that learn in isolation will be difficult to understand and control, while AI that is developed to join the human community of knowledge can accelerate the mechanisms that have made humanity such a successful species.



Here, we take a step toward reverse-engineering the cumulative cultural evolution process by studying multi-generational learning of novel environments in minimalist, Atari-style video games. 
Classic video games like Atari are rich test beds for studying adaptation to novel environments and thus cultural transmission \cite<cf.>{caldwell2008studying, mesoudi2008multiple}, as they involve learning many kinds of objects, complex dynamics, and adaptive behaviors, a feature which has attracted AI researchers in the past decade \cite<e.g.,>{guo2014deep, mnih2015human}.
Human adults learn these games rather quickly, especially in comparison to Deep Reinforcement Learning models \cite{tsividis2017a, lake2017building} in part due to richly structured prior knowledge about objects and physics \cite{dubey2018investigating}. 
Indeed, in such games, human learning can be further accelerated through verbal instruction \cite{tessler2017avoiding, tsividis2017a}, suggesting that language may be a robust mechanism of cultural knowledge transmission in flexibly solving novel tasks.
In these previous studies, however, the knowledge of a game that was communicated was a correct and complete description (e.g., the rule book or expert knowledge).
Language's ability to convey intermediate and incomplete stages of understanding a task -- a prerequisite for gradual accumulation of knowledge across generations -- has not been examined. 

As a first look into the cultural evolution process in novel video games, we design an iterated learning experiment where participants learn about new  games and share their knowledge with a subsequent generation of participants, who then learn from the previous generation and their own experience \cite{kalish2007iterated, beppu2009iterated}. 
We compare the knowledge and skill acquired by our multigenerational chains of participants with a control set of participants who learn by themselves.
We find that cultural learning mediated by language can be just as efficient as individual lifetime learning and follows a surprisingly similar trajectory.
We additionally find that the body of knowledge encoded in free-form linguistic messages increases over generations, enabled by high-performing individuals revising and expanding the cultural body of knowledge. 
These early results suggest that language is a powerful and flexible mechanism for transferring knowledge  about solving complex, novel tasks.

\vspace{-0.3cm}
\section{Human Experiment}

We designed an iterated learning experiment wherein participants play novel video games for a limited amount of time and then  share their knowledge with a future generation of participants in an open-ended linguistic message.
Data from this experiment and analysis scripts used to generate results and figures can be found at \url{https://github.com/jmadeano/growing_knowledge_culturally_neurips21}.
All data is anonymized, and anyone is free to share and adapt this data, provided credit is given (Creative Commons Attribution 4.0 International).

\begin{figure*}[t]
\includegraphics[width=\textwidth]{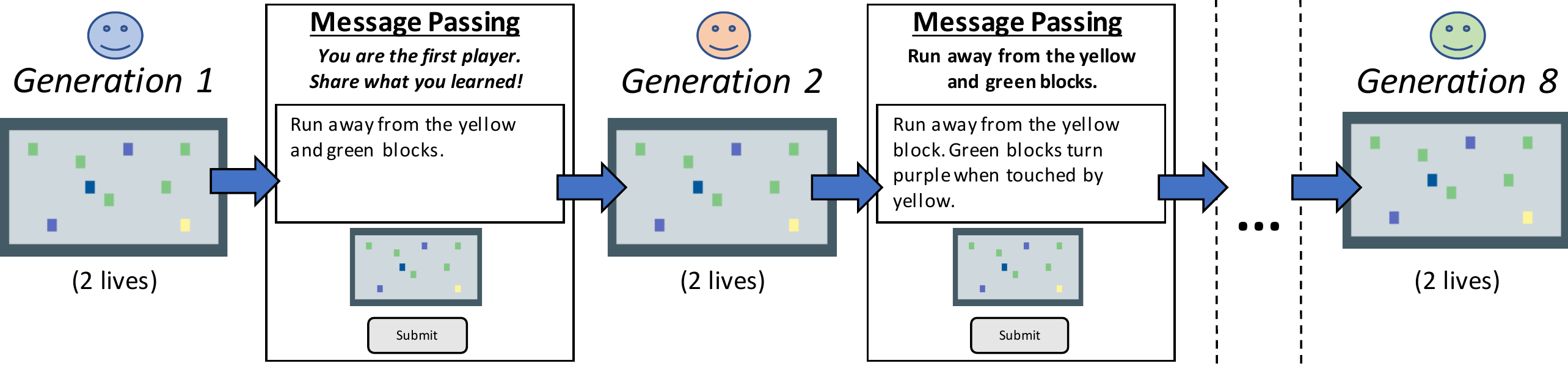}
\caption{Experiment flow for one game with a single chain of participants. Each generation is a different participant. The first generation learns from scratch to play the game, until their two lives are spent. They then pass a message to the next generation, and the process repeats. The text-box for message passing is initialized with the previous generation's message, which may be deleted, edited, or added to. Full study contained ten chains, which each ran for eight generations; each generation played ten different games.}
\label{fig:transmission_flow}
\vspace{-0.6cm}
\end{figure*}


\vspace{-0.12cm}
\subsection{Methods} 
\vspace{-0.08cm} 

\subsubsection{Procedure}

Participants ($n=80$, see Supplement for details) were organized into ten chains of eight participant-generations each (Figure \ref{fig:transmission_flow}). The participants read a set of instructions about the experiment and were asked a set of comprehension questions. Participants who failed to answer all questions after three attempts were not allowed to continue with the experiment. Each participant completed ten trials, which corresponded to playing ten different video games, each consisting of three phases: message reading, game play, and message passing (described below). 
Following the completion of all ten games, participants completed a demographic questionnaire which included a question about previous video game experience.
The only information about the games that the participant was told in the instructions was that the games used the arrow keys and some games used the space bar. All generations except the first were also told that they would receive a message from a previous player of the game.
Participants were told they would receive (and were, in fact, given) bonus pay based on both their performance and that of the next generation, thus providing an explicit incentive for helpful messages.

\noindent\textbf{Message reading}
Before playing each game, participants were provided a message about the game from a previous player of the game. Participants in the first generation were not provided any message.

\noindent\textbf{Game play}
Each game contained between four and six levels; different levels of a game operated according to the same general rules of the game, but varied according to the number or kinds of objects in play. Later levels of a game tended to be more challenging.
The game play screen displayed the game (trial) number, level number, number of additional lives remaining (1 or 0), and the text of the message that the player received. 

All generations of players started a game from the beginning (i.e., first level).
Participants clicked a `Start' button to start the game.
If the player beat the level, they proceeded to the next level of the game until there were no more levels to the game. 
Upon losing (dying), participants were able to restart from the last level they were playing.
For players who got stuck, a button appeared after 45 seconds of gameplay which allowed players to sacrifice a life in order to start the level over (or end the game, if no lives remained). 
The game ended when the participants lost twice (i.e., when both lives were used up).

\begin{wrapfigure}{r}{0.525\textwidth}
\begin{center}
\vspace{-0.6cm}

    \caption{Examples of two VGDL games. \textbf{Zelda} A: Monsters (pink, yellow, brown) kill avatar (blue) upon contact, but can be killed by the avatar by using the spacebar (``a sword'') when next to them. (B) Avatar must pick up the key (orange) before collecting the goal block (green). \textbf{Avoid George} A: Citizens (green) turn purple if George (yellow) touches them. Game is lost if all citizens turn purple or if George touches the avatar (blue). B: Avatar can revert citizens to green by using the spacebar when next to them. Game is won by keeping at least one citizen green for approximately twenty seconds.} \label{games_fig}
        \includegraphics[width=0.52\textwidth]{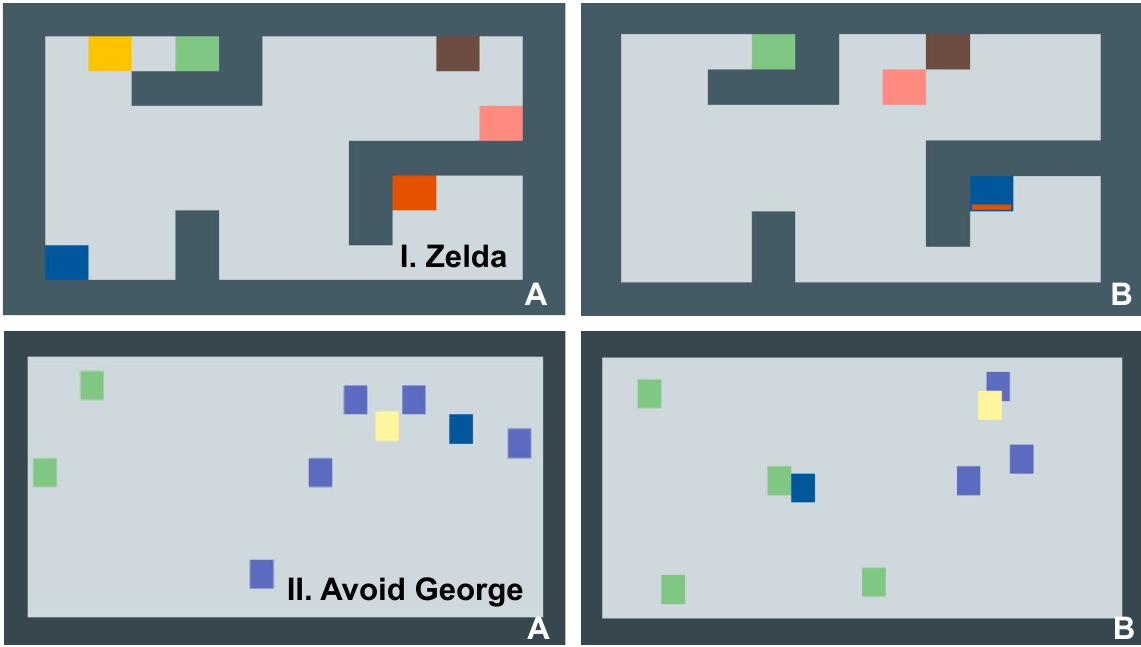}
\end{center}
\vspace{-1cm}
\end{wrapfigure}

Participants played two easy games in a fixed order (\emph{Explore/Exploit}, \emph{Preconditions}) in order to get acclimated to the controls and gameplay.
The order of the remaining eight games was randomly determined for each chain, but was constant for all generations of a chain.

\noindent\textbf{Message passing}
After either completing all levels of a game or expending their two lives, participants went to a message-passing screen (Fig.~\ref{fig:transmission_flow}). They were told that their message could contain any information that they thought would be helpful to a future player of the game and that they could use the message they received from the previous participant as the basis for their own message (i.e., they could delete, edit, or add to the last message). The text-box on the screen was initialized with the message they had received previously. After submitting their message, the participant proceeded to the message reading screen for the next game and repeated the process until they finished all the games.

\subsubsection{Materials}

The ten games in our study were inspired by classic Atari video games and encoded in a Video Game Description Language (VGDL) \cite{schaul2013video}. In all of the games, participants play as an avatar which they can move by using the arrow keys, and different object types manifest as blocks of different colors. In some games, the space bar can be used to interact with certain objects. 

The games are simple, but vary considerably in terms of their goals, dynamics, and winning strategies (see Supplement for brief descriptions). In \emph{Zelda}, for instance, the player must pick up a key and reach a door while avoiding monsters (Fig.~\ref{games_fig}); in \emph{Lemmings}, the player must clear a path to help lemmings reach an exit; in \emph{Sokoban}, the player must solve a puzzle by pushing boxes into holes. Most games are won when the player has picked up or destroyed all instances of some object type; in one game (\emph{AvoidGeorge}), the player wins by surviving until a timer runs out.

\subsection{Results}

Very few (6 of 80) participants reported having any experience playing these games before, and the experience reported was quite limited (the most experience reported was somebody having played two of these games before). We did not ask participants detailed questions about their video game experience in general. 


\vspace{-0.4cm}
\begin{wrapfigure}{l}{0.5\textwidth}
\begin{center}
\vspace{-0.7cm}
    \includegraphics[width=0.48\textwidth]{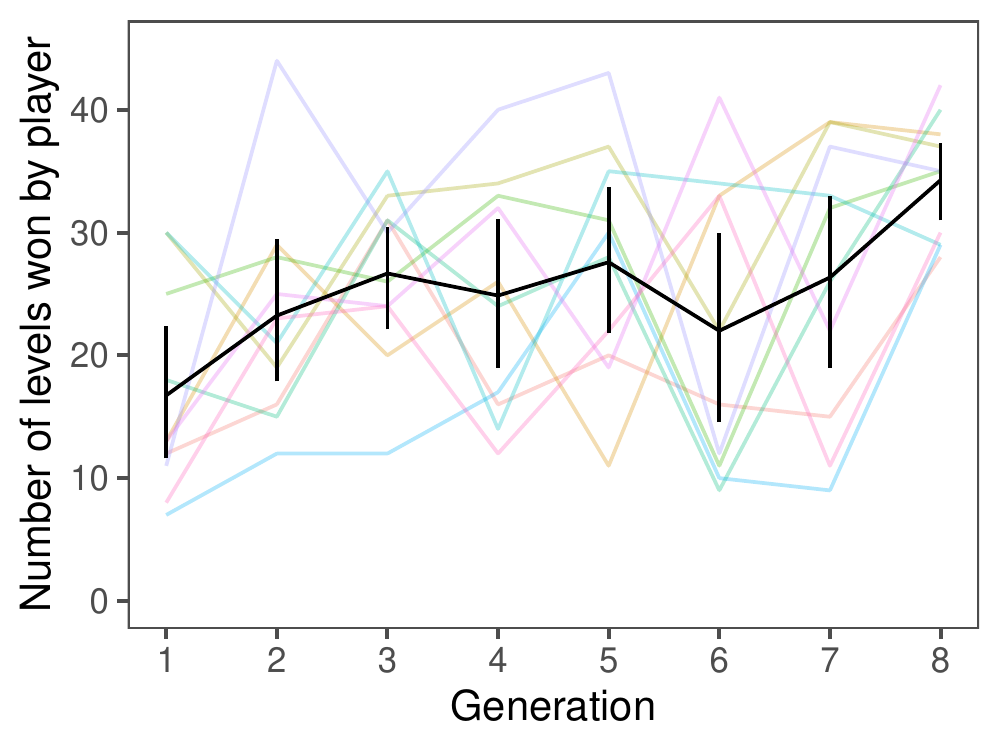}
\vspace{-0.3cm}
    \caption{Levels won by a player over the course of playing 10 novel games, as a function of the generation in a chain. Colored lines show different chains. Error-bars denote bootstrapped 95\% confidence intervals (CIs).}
    \label{fig:aggregateWins}
\end{center}
\vspace{-0.5cm}
\end{wrapfigure}

\paragraph{Mastery of the tasks}

Participants learned the basics of many of these games rather quickly. 
We observe that even given just two attempts at each game, the first generation of participants successfully completed almost 20 levels across the 10 game experiment. 
Notably, we observe that the ability to successfully complete a level of a game increased as the generation number increased (Fig.~\ref{fig:aggregateWins}), as confirmed by a mixed-effects model with random-effects by chain of intercept and generation slope ($\beta = 1.52$; $SE = 0.42$; $t = 3.6$).
This result is the first indication that language is a mechanism that can transfer actionable knowledge across generations to solve novel, complex tasks.

\begin{figure*}[t]
\includegraphics[width=1\textwidth]{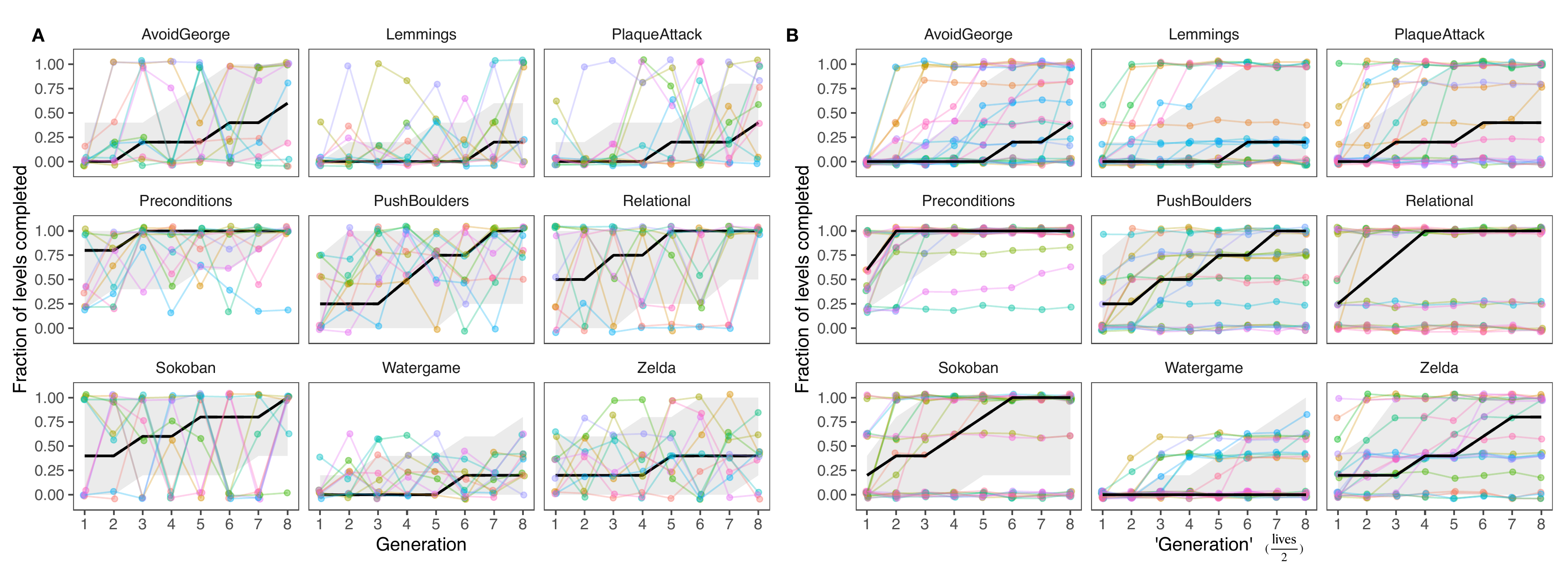}
\vspace{-0.6cm}
\caption{Learning (A) across generations and (B) within individuals for nine games, in terms of the fraction of a game completed (from 0\% to 100\%). Individuals in (A) played the each for 2 lives, before passing a message to the next generation. Individuals in (B) had an unlimited number of lives, and lives are chunked into `generations' (every 2 lives) to facilitate comparison across panels.  Colored lines represent different chains for (A) and different participants for (B). Solid black line is MAP estimate of the posterior predictive ordinal regression line; shaded area denotes 50\% Bayesian credible interval.}
\label{fig:chainsResults}
\vspace{-0.5cm}
\end{figure*}

The variability of the ten video games highlights the generality of language as a flexible means to communicate knowledge. 
The performance curves across generations for the individual games vary widely (Fig~\ref{fig:chainsResults}A).
To quantitatively model this variability, we built a Bayesian mixed-effects cumulative-logit (ordinal) regression model predicting the number of levels that a player completed as a function of their generation in the chain; in addition, we included random effects of intercept and slope (effect of generation) as a function of the unique chain and of the unique game.\footnote{
The lmer-style model code is 
 $\texttt{levels\_won} \sim \texttt{generation} + 
  (1 + \texttt{generation} | \texttt{chain}) +
  (1 + \texttt{generation} | \texttt{game})$
.
We omit from this analysis the game \emph{ExploreExploit};  there is no way to die in this game and all players completed all levels.
 }
 We observed a small, positive effect of generation on a player's performance (number of game levels won, $\beta =  0.21$; 95\% Bayesian credible interval $(0.11, 0.31)$). 
 In addition, we observe that the variability of the slope of generation as a function game was relatively high ($\sigma^{game}_{gen} = 0.08$), about a third of the size of effect of the generation and higher than the variability by chain ($\sigma^{chain}_{gen} = 0.05$).
 We also observe dramatic swings for certain games in the level achieved from generation to generation within a chain (e.g., in \emph{Sokoban}); this within-game variability likely reflects participants' variable baseline video-game skill, especially for the games that require dexterous use of the keyboard. 
Still, due to our living document paradigm, the content of a message often persists across players who struggle;  hence, chains that collapsed (e.g., a participant failing to beat even the first level, following a participant who beat the entire game) were often quick to recover (i.e., the next generation could completely master the game), underscoring the robustness that comes with accumulated cultural knowledge. 
 
\begin{wrapfigure}{r}{0.55\textwidth}
\begin{center}
\vspace{-0.5cm}
    \includegraphics[width=0.54\textwidth]{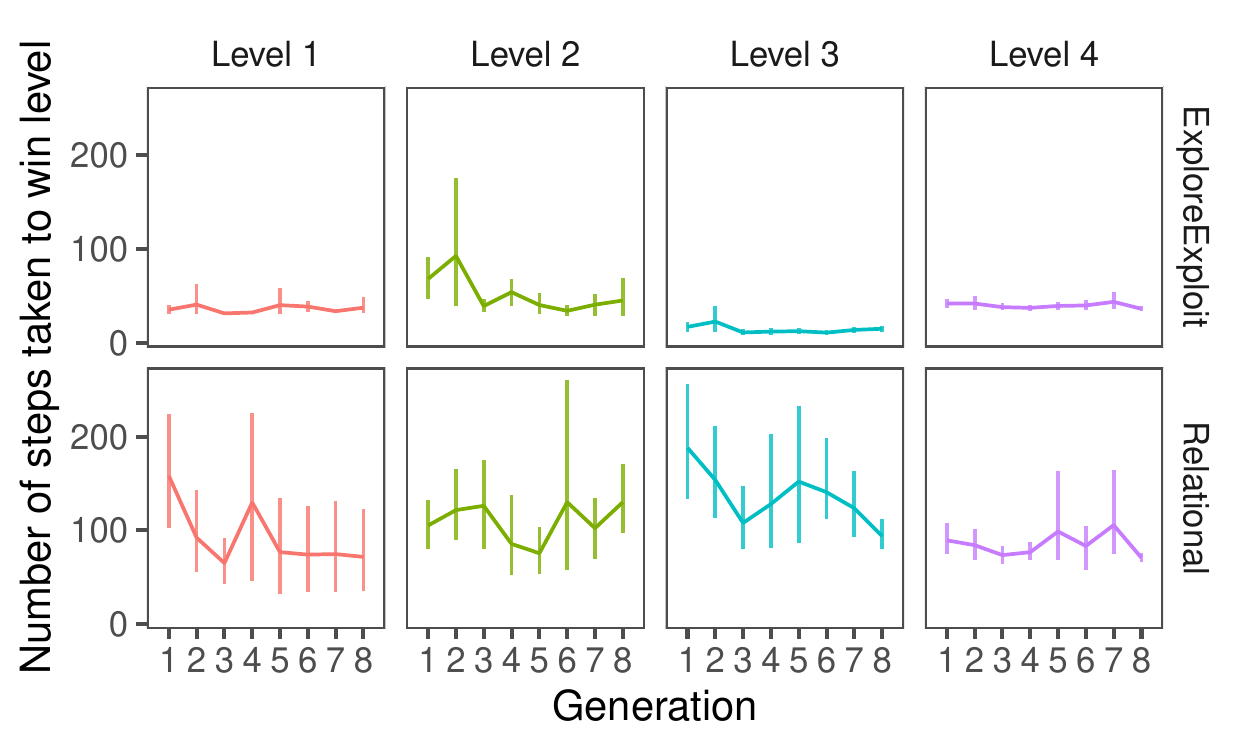}
\vspace{-0.3cm}
    \caption{Numbers of steps (key-presses) taken by players to complete each level of two different games as a function of their position in the transmission chain. Participants overall take fewer steps to successfully complete a level as the chain progresses. Error-bars denote bootstrapped 95\% CIs.}
    \label{fig:stepsByLevel}
\end{center}
\vspace{-0.5cm}
\end{wrapfigure}

\paragraph{Efficiency in play}
The knowledge that one acquires by learning from others not only allows one to survive and achieve one's goals in a novel environment, but it may also enable more efficient planning and acting in that environment. As a first-pass at examining efficiency in participants' actions, we look at how many steps, or key-presses, participants took to successfully complete a level of a game, as a function of their position in the generational chain.
We fit a Bayesian mixed-effects linear regression model predicting the number of steps a player took to successfully complete a level as a function of generation, with random intercepts and slopes by chain and by each level of each game. 
We find an appreciable decline in the number of steps needed to successfully win a level of a game: $\beta = -3.86 (-7.26, -0.44)$, suggesting that participants not only completed more of the task in future generations, but did so more efficiently (see Fig. \ref{fig:stepsByLevel} for two example games).




\vspace{-0.3cm}

\paragraph{Comparison to individual learning}

Participants in later generations of the chains both performed better at the task and achieved their successes more efficiently.
How does learning from the cultural knowledge accumulated over generations compare to learning from more substantial direct experience in the environment?
To begin to address this question, we compare our transmission chains of learners to a separate group of participants who learned these games by themselves with an unlimited number of lives \cite{tsividis2021}. 
As with the transmission chain participants, prior to playing the games, participants were told only that they could use the arrow keys and the space bar, and that they should try to figure out how each game worked.
Participants had the option to restart a given level if they were stuck, or to forfeit the rest of the game (and any potential earned bonus) if they failed to make progress after several minutes. 
Each participant played their games in a random order.

The learning trajectories exhibited across generations and those reached by individuals in our control experiment exhibit surprising similarities.\footnote{
The largest dissimilarity between the learning trajectories of individuals~vs.~chains is the variance. Chains exhibited greater variance than individuals because each generation of a chain re-started the game from the game's first level, whereas individuals always restarted the game from the last level they reached.} In most games, the average transmission chain reaches the same maximum level as the average isolated individual. Transmission chains and isolated individuals also reached their maxima in roughly the same number of lives  (Figure \ref{fig:chainsResults}B shows performance in terms of number of lives, chunked into ``2-life generations'' for easy comparison to the chains). 
Most strikingly, each game appears to have a characteristic signature learning curve: the best visual match for the learning curve of each game in the transmission setting is usually the game of the same name from the individual condition.
Indeed, the area under the learning curves for each game show a strong correlation between the individual- and transmission settings ($r = 0.87$).
This result suggests that the challenges faced in each game by a hypothetical, long-lived player are recapitulated across successive generations of individual players.

\begin{figure}[h]
\vspace{-.2cm}
   \begin{center}
   \begin{tabular}{|p{.95\linewidth}|}
  \hline
  \setul{1ex}{.3ex}
  \setstcolor{Blue} 
\noindent\scriptsize\texttt{\textcolor{Purple}{You control the white block.} Push \st{all the blocks into each other.} \textcolor{Blue}{the blue blocks into the yellow blocks. If there are no yellow blocks, you can create them by \setstcolor{Red} \st{pushing red blocks into blocks of other colors} \textcolor{red}{simply pushing red blocks. Yellow blocks can also be created by pushing blocks of other colors into each other} (such as purple or pink), and then push the blue blocks into the yellow blocks you've created.} \textcolor{red}{You cannot move purple blocks.} \setstcolor{Green}\textcolor{Purple}{Be careful not to push the block into a corner.}} \\ \hline
\end{tabular}
\vspace{-0.2cm}
\end{center} 
    \caption{Example of message dynamics across generations, from players describing the game  \emph{Relational}. Text contributions throughout generation 1-4 are shown in black, blue, red, and purple, respectively.  Some participants made significant rewrites as they learned about the games while others made little to no change to the message they received.}
    \label{fig:messageExample}
\end{figure}

\vspace{-0.3cm}
\begin{wrapfigure}{r}{0.55\textwidth}
\begin{center}
\vspace{-0.5cm}
    \includegraphics[width=0.54\textwidth]{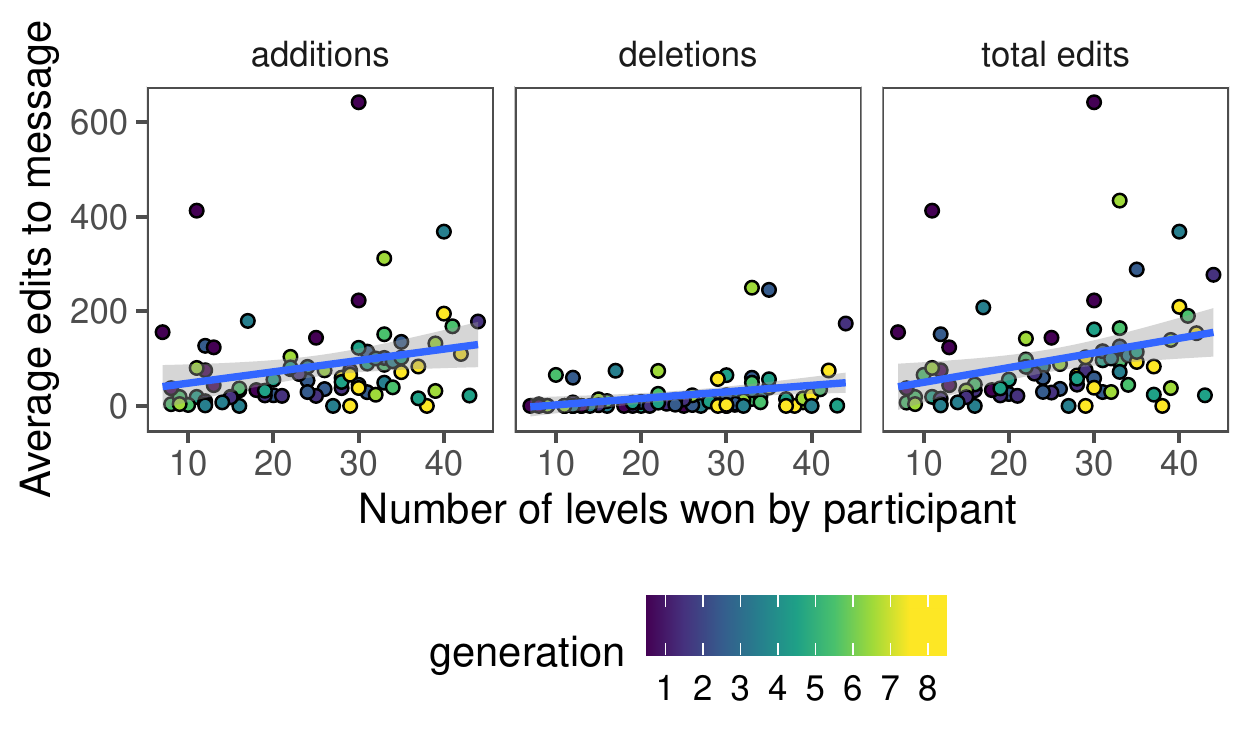}
\vspace{-0.3cm}
    \caption{Amount of change to the message document (additions, deletions, total edits) as a function of the success of the sender. Participants who succeeded at more of the task made more changes, both in terms of additions and deletions. Points represent participant averages over the ten games.}
    \label{fig:editDistance}
\end{center}
\vspace{-0.5cm}
\end{wrapfigure}
\paragraph{Message Forms}
The messages that are passed from generation to generation are a kind of living document that has a life and evolution of its own (Fig.~\ref{fig:messageExample}). 
As a first pass, we used Python NLTK to tokenize each of our 800 messages into 3994 component sentences and to compute the Levenshtein edit distance between messages of consecutive generations \cite{nltk}.
The number of sentences $n$ that made up a message increased across generations at a rate of approximately 0.5 sentences per generation (median $n_{gen1} = 2$, $n_{gen8} = 6$). 
Participants not only added to the document, they also removed content from the written record. 
On a character-by-character analysis, approximately 80\% of the edits were additions and 20\% were deletions, and the relative proportion remained constant across generations.
Intriguingly, the amount of editing of a message document varied as a function of the performance of the sender, with participants who performed well making more edits to the document than participants who succeeded less at the task (Fig.~\ref{fig:editDistance}). 
This relationship between sender performance and edits is present even when controlling for the effect of generation (as later generations tend to perform better); in a Bayesian mixed-effects regression model with fixed-effects of generation and performance (and with maximal by-game and by-chain random effects), we find that edits increase as a function of performance ($\beta = 67.17 [26.8, 109.1]$) though not necessarily as a function of generation ($\beta = -9.47 [-25.2, 6.3]$).
The fact that edits may slightly decrease across generations may be indicative of knowledge growth asymptoting; as the inherited body of knowledge becomes more refined and accurate, there is less of a need for major revisions. 



\paragraph{Message Contents}

The content of the linguistic messages provides a window into how cultural knowledge grows across generations.
We see that players talk freely about the game dynamics (``yellow squares can be reverted''), the goals of a task (``reach the green square''), as well as higher-order strategies for the avatar (``avoid trapping yourself'') and the human controller (``keep calm''). 
We developed a scheme to code whether each sentence contained information about game \emph{dynamics}, player \emph{policy}, and \emph{goals} (win conditions, loss conditions, or neither). Additionally, we coded the abstractness of the language used (\emph{concrete} if the sentence referred to objects/actions easily expressible in VGDL, and \emph{abstract} if not). Finally, a sentence was coded as \emph{ignorance}\slash\emph{experience} if it  expressed lack of knowledge or described specific experiences of the agent (see Supplement for full details on coding scheme). 
To label the full data set of sentences, we used the Elicit tool developed by Ought (\texttt{ought.org}) that uses the few-shot learning capabilities of the large language model GPT-3 \cite{brown2020language} to automatically code our corpus of natural language sentences \cite{elicit}.\footnote{
 For each tag (dynamics, policy, abstract, or valence), Elicit used the Curie GPT-3 model to check whether the meaning of the statement fit the tag; Elicit put labeled training examples into the prompt to show Curie the meaning of the tag.
 This system achieved 78.5\% cross-validation accuracy (Cohen's Kappa = 0.75) on our 200 human-labeled ground-truth sentences. 
}

\begin{figure}
\begin{minipage}[c]{0.58\textwidth}
    \includegraphics[width=1\textwidth]{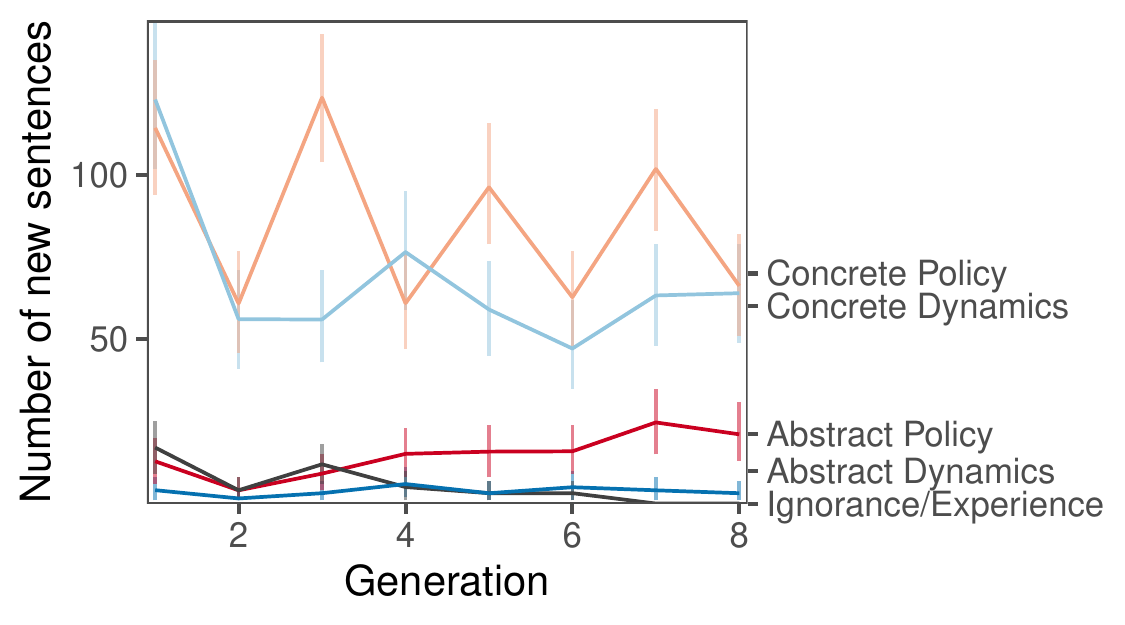}
\end{minipage}
\begin{minipage}[c]{0.42\textwidth}
    \includegraphics[width=1\textwidth]{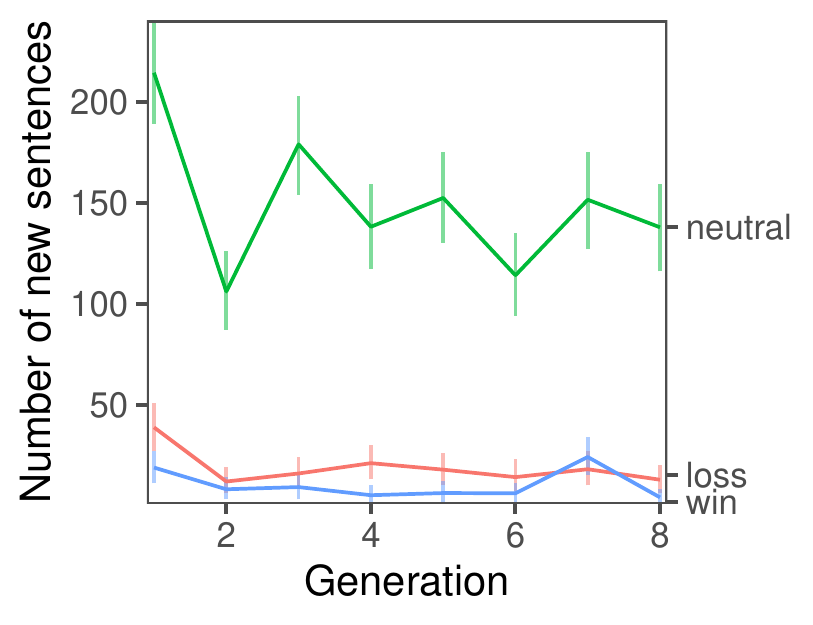}
\end{minipage}
    \caption{Content of messages (in number of sentences) that was added at each generation according to whether and how they referred to policy~vs.~dynamics information (left) and explicit reward conditions (right). See supplement for full details of coding scheme.}
    \label{fig:messageContentUniqueProp}
\end{figure}

The majority of sentences referred to either Concrete Dynamics or Concrete Policy information, in roughly equal proportions. We see that the amount of new Concrete Policy and Concrete Dynamics statements decreased substantially from the first to the last generation, Ignorance\slash Experience statements decreased over generations, while Abstract Policy statements tended to increase over generations (Fig. \ref{fig:messageContentUniqueProp} left). The vast majority of statements did not refer explicitly to positive or negative rewards, though we observed that negative rewards (loss conditions) were mentioned more than positive rewards (win conditions), and that both positive and negative reward statements appeared most often in the first generation and tapered off in subsequent generations (Fig. \ref{fig:messageContentUniqueProp} right).
Together, these results point to a natural curriculum that develops over generations: indicate what to avoid and what to approach (goals), describe how the world works and what actions to take (concrete dynamics, concrete policy), and then articulate more abstract strategies that will help you get to where you want to go (abstract policy).

\paragraph{Language and performance}
Does the success of a player impact the kind of language they send to the next generation? And, how does the language received impact the performance of the next generation?
To assess the first question, we ran independent Bernoulli mixed-effects models to predict the presence or absence of each of four sentence types as determined by Elicit (omitting Abstract Dynamics due to lack of sufficient data) as a function of the performance and generation of the message sender.\footnote{
The lmer-style model code is 
 $\texttt{sentence\_type\_produced} \sim  \texttt{generation} + \texttt{levels\_won} +
  (\texttt{generation} + \texttt{levels\_won}  | \texttt{game}) + 
    (\texttt{generation} + \texttt{levels\_won}  | \texttt{chain})$
 .
} 
We found, controlling for generation, high-performing players tended to produce more statements of Concrete Dynamics ($\beta = 0.77 [0.04, 1.54]$) and Concrete Policy ($\beta = 1.33 [0.64, 2.08]$), and avoided statements of Ignorance\slash Experience ($\beta = -2.31 [-4.22, -0.69]$); there was suggestive evidence that Abstract Policy statements were affected both by the performance of the player ($\beta =  0.68 [-0.19, 1.55]$) and by their generation ($\beta = 0.19  [-0.03, 0.41]$).
To assess the impact of the the message on the receiver, we examined the variability in overall learning of the games among chains by constructing a Bayesian mixed-effects regression model predicting the fraction of a game completed as a function of the language conveyed about that game by the previous generation in the chain. 
We used the Elicit tags to construct fixed-effects of five categories of linguistic messages (concrete policy, concrete dynamics, abstract policy, abstract dynamics, and statements of ignorance), in addition to fixed-effect of overall length (in characters) and the maximal random-effects by game.\footnote{
The lmer-style model code is 
 $\texttt{levels\_won} \sim -1 + \texttt{concrete\_dynamics} + \texttt{concrete\_policy} + \texttt{abstract\_dynamics} + \texttt{abstract\_policy} + \texttt{ignorance} + \texttt{length} +
  (-1 + \texttt{concrete\_dynamics} + \texttt{concrete\_policy} + \texttt{abstract\_dynamics} + \texttt{abstract\_policy} + \texttt{ignorance}  | \texttt{game})$.  We omit random effects of \texttt{chain} as this is the primary source of variability in \texttt{levels\_won} that we aim to explain.
} Concrete Dynamics and Concrete Policy showed the largest effects on player performance ($\beta_{dynamics}^{concrete} = 0.78 (0.52, 1.04)$; $\beta_{policy}^{concrete} = 0.87 (0.58, 1.15)$) and Abstract Policy information also showed an appreciable effect on performance ($\beta_{policy}^{abstract} = 0.50 (0.02, 0.99)$). 
There were very few Abstract Dynamics statements, and hence, the uncertainty around the effect of Abstract Dynamics information was too large to consider it appreciably effective ($\beta_{dynamics}^{abstract} = 0.96 (-0.20, 2.08)$); information that conveyed Ignorance\slash Experience was also not obviously helpful for the next generation ($\beta_{ignorance} = 0.20 (-0.93, 1.22)$); there was no effect of overall length of the message above the beyond these sentence types ($\beta_{length} = 0 (0, 0)$).


\vspace{-0.1cm}
\section{Discussion}

The ingenuity of a human being is the byproduct of accumulated learning over a time scale that extends beyond any one lifespan. 
Here, we take a first step at reverse-engineering the cumulative cultural learning process in humans by investigating transmission chains of individual learners faced with novel, complex tasks in the form of minimalist video games. 
We find that, even with a limited lifespan (``you only live twice''), our participants learned enough and transmitted a sufficiently faithful representation of their knowledge to enable the subsequent generations to pick up where they left off. 
In doing so, we find cultural learning mediated by language to be just as efficient as individual lifetime learning and to follow a surprisingly similar trajectory to individual learning. 
In this way, cultural learning allows individual learners to get much further in their brief lives than learners who have to start from scratch.
They do this by conveying distinct kinds of information encoded flexibly in natural language: Dynamics about how the world works, information about what goals are worth pursuing and what you should avoid, strategies about how to succeed.
Furthermore, learners encode increasingly abstract information over generations, suggesting that there is a natural curriculum to pedagogical teaching over generations. 
Finally, these specific aspects of knowledge encoded in language are associated with the success of chains, above and beyond the undifferentiated quantity of raw content of the messages.

Building machines that can learn and teach as flexibly as humans is a crucial step on the path to Cooperative AI \cite{dafoe2020open}.
Examples of extant approaches towards this goal have taken the form of knowledge distillation (e.g., from one large neural network to a smaller neural network \citeNP{yim2017gift}) and text summarization (e.g., \citeNP{nallapati2016abstractive}).
To gain the most flexible communicative agents, AI systems should incorporate human participants into their learning process, both to learn from humans effectively and efficiently and to be able to transmit learned knowledge back to humans.

  

In our experiment, we employed a ``living document'' design where the message of one generation was seeded with the (editable) text of the previous generation. 
This mode of transmission is one of the richest forms of linguistic transmission, as the mechanism itself provides a kind of cultural ratchet.
We cannot guarantee that other forms of linguistic transmission (e.g., single, unseeded messages) would or would not result in similar knowledge accumulation we observed \cite<cf.,>{beppu2009iterated}.
It is also possible that language is not necessary for cumulative cultural learning in our setting, as suggested by results in other tasks \cite<e.g.,>{caldwell2008studying}; however, task difficulty may be an important third variable when evaluating the necessity of language for cultural learning \cite{Morgan2015}.
Our video game paradigm is natural to extend to other modalities of transmission, including nonlinguistic pedagogical demonstration and incidental observational learning (e.g., through video-game replay). 

Our comparisons to the participants learning by themselves also should be contextualized. These control participants were endowed with an unlimited number of lives (or attempts) in the game, which may have invoked a different trade-off in explore~vs.~exploit behavior. 
This change might have resulted in different, finer-grained individual behavior, though the overall time-courses (as measured in lives) of the trajectories remained remarkably consistent. 
The explore~vs.~exploit trade-off may further be modulated by the conditions one assumes you need to satisfy to send a message; here, participants had the luxury of being able to live (twice) and then send a message. 
When the transmission of the message itself depends upon one's survival, complex dynamics may arise.

Much work in cumulative cultural evolution examines learning how to engineer and use tools \cite<e.g.,>{derex2019causal, osiurak2021technical}, but tool knowledge is just one part of the cognitive repertoire that allows our species to flexibly adapt to solve complex, novel tasks:
We must also understand cause--effect relationships, the dynamics of the world, what goals are worth pursuing, what pitfalls are likely to emerge, abstract strategies, and more. 
One particularly powerful aspect of human language is that it enables the sharing of abstract knowledge that goes beyond direct experience \cite{koenig2004trust, gelman2009learning, harari2014sapiens}.
Understanding the power of language for cultural transmission will be key to unpacking how the life of a human seems to extend in all directions, time and space.

\section*{Acknowledgments}

This material is based upon work supported by the National Science Foundation SBE Postdoctoral Research Fellowship Grant No. 1911790 awarded to M.H.T. and Army Research Office MURI Grant No. W911NF-19-1-0057.

\newpage
\section*{Checklist}
\begin{enumerate}

\item For all authors...
\begin{enumerate}
  \item Do the main claims made in the abstract and introduction accurately reflect the paper's contributions and scope?
    \answerYes{}
  \item Did you describe the limitations of your work?
    \answerYes{We discuss how our experimental setup constitutes just one point in a vast design space of experiments. See the discussion section.}
    \item Did you discuss any potential negative societal impacts of your work?
    \answerNA{We are not aware of any immediate potential negative societal impacts of the work, and we believe that more AI systems incorporating human participants into their learning process will be beneficial (provided consent is received and human data is unbiased/representative).}
  \item Have you read the ethics review guidelines and ensured that your paper conforms to them?
    \answerYes{}
\end{enumerate}

\item If you are including theoretical results...
\begin{enumerate}
  \item Did you state the full set of assumptions of all theoretical results?
    \answerNA{}
	\item Did you include complete proofs of all theoretical results?
    \answerNA{}
\end{enumerate}

\item If you ran experiments...
\begin{enumerate}
  \item Did you include the code, data, and instructions needed to reproduce the main experimental results (either in the supplemental material or as a URL)?
    \answerYes{All the (anonymized) participant data is available in a linked GitHub repository along with the code necessary to generate all figures in the paper. The repository has been anonymized for the initial submission.}
  \item Did you specify all the training details (e.g., data splits, hyperparameters, how they were chosen)?
    \answerNA{}
	\item Did you report error bars (e.g., with respect to the random seed after running experiments multiple times)?
    \answerYes{See all figures.}
	\item Did you include the total amount of compute and the type of resources used (e.g., type of GPUs, internal cluster, or cloud provider)?
    \answerNA{}
\end{enumerate}

\item If you are using existing assets (e.g., code, data, models) or curating/releasing new assets...
\begin{enumerate}
  \item If your work uses existing assets, did you cite the creators?
    \answerYes{Cited data used from \cite{tsividis2021}}
  \item Did you mention the license of the assets?
    \answerYes{See the Human Experiment section.}
  \item Did you include any new assets either in the supplemental material or as a URL?
    \answerYes{All analysis code and data are available in a companion repository linked in the paper. See the Human Experiment section.}
  \item Did you discuss whether and how consent was obtained from people whose data you're using/curating?
    \answerYes{Consent was obtained from the authors of \cite{tsividis2021}.}
  \item Did you discuss whether the data you are using/curating contains personally identifiable information or offensive content?
    \answerYes{All participant data is anonymized. See the Human Experiment section.}
\end{enumerate}

\item If you used crowdsourcing or conducted research with human subjects...
\begin{enumerate}
  \item Did you include the full text of instructions given to participants and screenshots, if applicable?
    \answerYes{A description of the task and instructions are provided in the Human Experiment section, Figure \ref{fig:transmission_flow}, and in the Supplement. More detailed instruction and screenshots are provided in the supplement.}
  \item Did you describe any potential participant risks, with links to Institutional Review Board (IRB) approvals, if applicable?
    \answerNA{Exempt -- minimal risk experiment}
  \item Did you include the estimated hourly wage paid to participants and the total amount spent on participant compensation?
    \answerYes{Experiment completion time varied per participant, but we provided the median completion time and the base pay. See the Human Experiment section in the supplement.}
\end{enumerate}

\end{enumerate}


\bibliographystyle{apacite}

\bibliography{main}

\end{document}


\maketitle


\appendix

\section{Human participants}

We recruited 80 participants from the crowd-sourcing platform Prolific. Participants were restricted to those located in the US or Canada with at least a 95\% work approval rating, at least 5 previously completed studies, and English as their first language. An additional 38 participants were recruited to pilot earlier versions of the experiment; these participants were prevented from completing the final study.
Participants in the first five generations were paid \$7.35 as base pay and had a median completion time of 47 minutes. Later generations were paid \$9.20 as base pay as the median completion time increased to 57 minutes. All participants had the opportunity to earn up a bonus of up to \$2.00 based on a combination of how well they and the person who received their message performed.

We additionally compare our results to a subset of a previously collected data set of 300 participants recruited through Amazon Mechanical Turk (Tsividis et al., 2021). Participants in this experiment were paid \$3.50 plus a bonus of up to \$1.00 depending on their cumulative performance across all the games they played. 
\section{Games used in study}

Participants in our experiment played each of the following ten games, in a semi-randomized order. Participants always played Explore/Exploit first and Preconditions second; the remaining eight games were played in an order that was randomized for each chain (but constant within a chain).

\begin{table*}[!h]
\begin{center} 
\label{game-descriptions} 
\vskip 0.12in
\scriptsize
\begin{tabular}{lp{0.53\linewidth}p{0.32\linewidth}}
 \hline
\textbf{Game} &
  \textbf{Description} &
  \textbf{Key Features} \\ \hline
Explore/Exploit &
  Collect all blocks of any color. &
  Easy game; No way to die; multiple ways to win; time incentive \\ \hline
Preconditions &
  Collect a gem which is surrounded by poison. Poison can be passed through only after drinking an antidote. &
  Easy game; relevance of preconditions (sub-goals) \\ \hline
Zelda &
  Collect a key and then reach a door. Monster blocks try to kill you. Monsters can be destroyed using the spacebar. &
  Preconditions; use of spacebar \\ \hline
Plaqueattack &
Shoot ``toothpaste'' to destroy ``food''. Clone of the classic \emph{Spaceinvaders}. &
  Speed is important; use of spacebar \\ \hline
Sokoban &
 Puzzle game: Push boxes into holes. &
  Complex strategic planning; dead-ends; time incentive \\ \hline
Avoid George &
  George moves around turning green blocks into purple.  Avatar can revert this using the spacebar. Player wins if a green block remains when timer finishes. &
  Defensive play; use of spacebar \\ \hline
Push Boulders &
  Get to the exit by pushing boulders to clear dangerous obstacles. &
  Combination of strategic planning \& knowledge about object interactions \\ \hline
Relational &
  Make all blue gems disappear by pushing them into fire. Blocks may be converted into other block types and/or into fire. &
  Requires a large set of knowledge about object interactions; time incentive \\ \hline
Watergame &
  Puzzle game: Get to the exit, which is usually surrounded by deadly water that be cleared using dirt. &
  Complex strategic planning; dead-ends \\ \hline
Lemmings &
  Help a group of lemmings get to their own exit. To do so, use spacebar to shovel a tunnel. &
  Player actions are costly (score-wise) but necessary; use of spacebar
\end{tabular}
\end{center} 
\vspace{-0.3cm}
\small{\caption{Overview of the ten games used in the experiment.}}
\vspace{-0.5cm}
\end{table*}

\begin{figure*}[tbh]
\includegraphics[width=\textwidth]{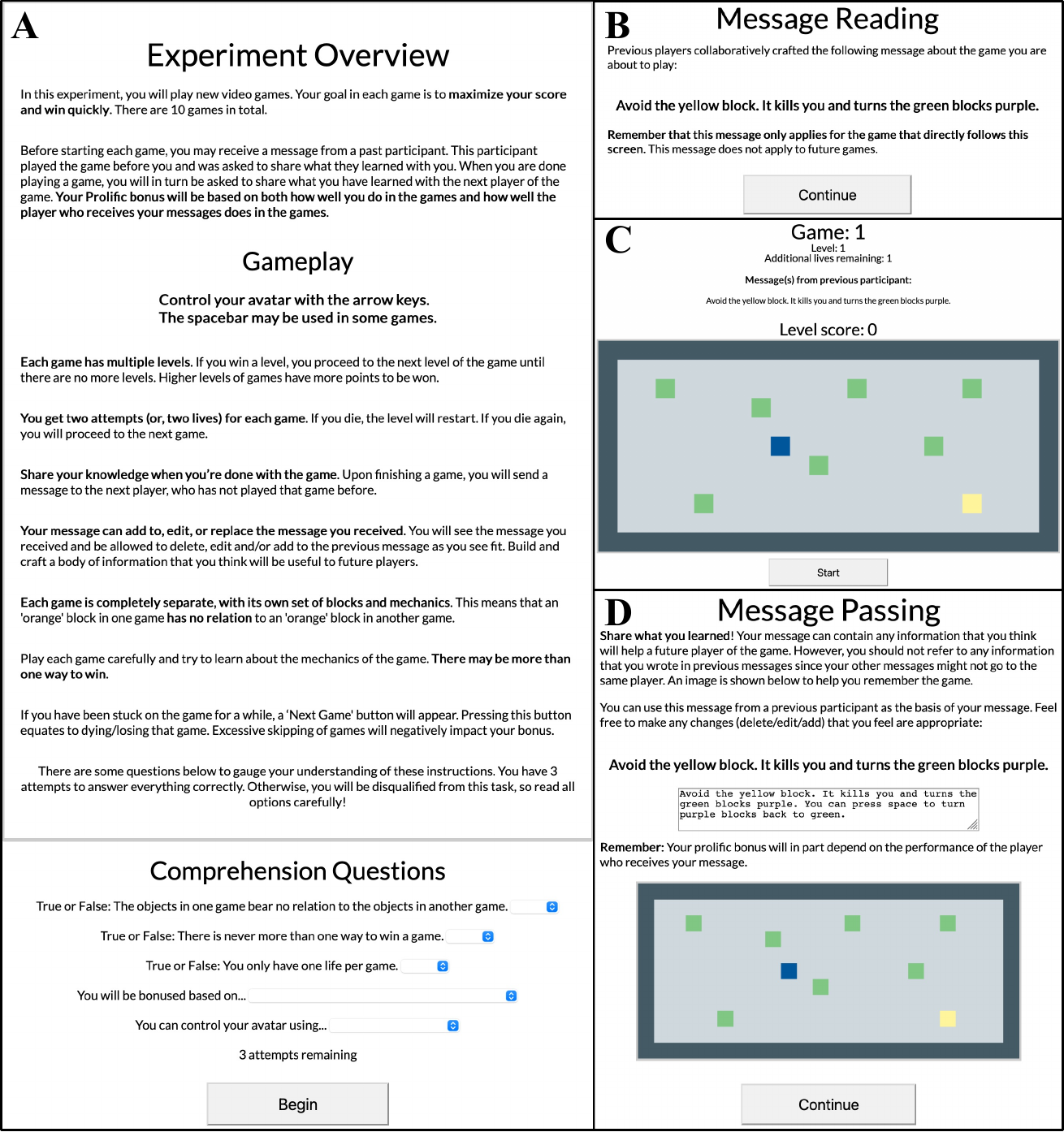}
\caption{Screenshots of each stage of the experiment. A) Participants receive an overview of the experiment and information about game controls and bonus opportunities. B) Participants are able to read the message left for them by a previous participant. C) Participants play the game. D) Participants are reminded of the message they received, provided with a screenshot of the start of the level to aid in remembering colors/dynamics, and asked to add/edit/delete from the message they received to craft a new message for the next participant. Participants repeat steps B-D for all ten games.}
\label{fig:transmission_flow}
\vspace{-0.6cm}
\end{figure*}

\section{Language coding}
The language coding scheme was developed iteratively until high agreement was reached on all categories between two human coders on a subset of unique 200 sentences. The coding scheme was then condensed into simple but descriptive phrases, so that the sentences could be classified by the Elicit tool (underpinned by GPT-3). The prompt provided to GPT-3 began with ``Classify the following as one of:'' and then gave the tag options; the prompt then continued with human-annotated examples in a list of ``sentence:'' and ``Tag:''.  The API responds with the most likely (highest log probability) tag among the options. 

While condensing our coding scheme into tags for Elicit, we explored many possibilities. For boolean features such as dynamics, we tried boolean tags (``true'', ``false'') as well as short, minimal tags (``dynamics'', ``not dynamics''). We found that choosing tags with succinct, natural language descriptions (``dynamics, or how the world works including explanations or affordances'') led to significant performance increases. All the final tags for each feature are listed below.

\paragraph{Dynamics}
Dynamics is a boolean feature, and the Elicit tool selected the most likely tag among ``dynamics, or how the world works including explanations or affordances'' and ``not dynamics, or how the world works including explanations or affordances''. 

\paragraph{Policy}
Policy is a boolean feature, and the Elicit tool selected the most likely tag among ``policy, or what actions to take including strategies or instructions'' and ``not policy, or what actions to take including strategies or instructions''.

\paragraph{Abstractness}
Abstractness was broken into three categories (``abstract, complex, high-level information'', ``concrete, simple, low-level information'', and ``ignorance statements or specific experiences''). 

\paragraph{Valence}
 Valence meant to pick out whether sentences contain any positive or negative notions of reward. The Elicit tool selected the most likely tag among ``losing, including information about death, losing points, lowering scores, forfeiting, losing lives, getting stuck or trapped'', ``winning, including mentions of scoring points, victory, success, goals, solutions, best strategies'', and ``neutral information''
 
We made a composite classification scheme to simplify the analyses. First, we turned the Ignorance level of abstractness (ignorance statements or specific experiences) into its own category, disregarding whether or not it had policy or dynamics information. Second, we prioritized policy information such that if a statement contained both policy and dynamics information, we classified it as Policy. We also did not examine the Valence categorized in its intersection with the other categories. Thus, for our analyses, we used a classification scheme of five, mutually exclusive categories for each sentence: Concrete Dynamics, Concrete Policy, Abstract Dynamics, Abstract Policy, and Ignorance\slash Experience.
 
\begin{table*}
\begin{center} 
\label{language-coding} 
\vskip 0.12in
\scriptsize
\begin{tabular}{C{0.10\linewidth}p{0.46\linewidth}C{0.06\linewidth}C{0.06\linewidth}C{0.08\linewidth}C{0.08\linewidth}C{0.05\linewidth}C{0.05\linewidth}}
\toprule 
            Game &                                                                                                                                                                                              Sentence & Abstract Policy & Concrete Policy & Abstract Dynamics & Concrete Dynamics & Ignorance & Valence \\ \hline

           Zelda &                                                                                                                             Don't go too fast when close to the moving squares, they're very erratic. &               X &                 &                 X &                   &           &         \\ \cline{2-8}
            &                                                                                                                                                                Carry the red block to the green goal. &                 &               X &                   &                   &           &       + \\ \cline{2-8}
            &                                                                                                                                The other cubes will be moving quickly and randomly and will harm you. &                 &                 &                 X &                   &           &       - \\ \cline{2-8}
            &                                                                                                                         The moving squares are very fast and do not seem to move in a logical manner. &                 &                 &                 X &                   &           &         \\ \cline{2-8}
            &                                                                                                                                                                                Every movement counts. &                 &                 &                 X &                   &           &         \\ \cline{2-8}
            &                                                                                                                                                       You die if you touch one of the moving squares. &                 &                 &                   &                 X &           &       - \\ \cline{2-8}
            &                                                                                                                                                                           Not sure, was unsuccessful. &                 &                 &                   &                   &         X &       - \\ \hline
         Sokoban &                                                                                                                                          Be careful to not push a green block where it can get stuck. &               X &                 &                 X &                   &           &       - \\ \cline{2-8}
          &  Pushing the green squares into the bottom red square is the safest route because there is more open space on the bottom half of the map and harder to get a green square stuck if you play it right. &               X &                 &                 X &                   &           &         \\ \cline{2-8}
          &                                                                                                                                                                     Ignore the red square at the top. &               X &                 &                   &                   &           &         \\ \cline{2-8}
          &                                                                                                                 You can go through the red squares freely without losing the game or getting blocked. &                 &                 &                 X &                   &           &       - \\ \cline{2-8}
          &                                                                                                                                                                                You can move the green &                 &                 &                   &                 X &           &         \\ \cline{2-8}
          &                                                                                                                                                                           You can go through the red. &                 &                 &                   &                 X &           &         \\ \hline
      Relational &                                                             Never push a movable box to the outside perimeter of the game field or it will get stuck there, and cannot be used to complete the level. &               X &                 &                 X &                   &           &       - \\ \cline{2-8}
       &                                                       Do not get any squares stuck into a corner or a wall because there will be no way to push them back out and you will have to forfeit the level. &               X &                 &                 X &                   &           &       - \\ \cline{2-8}
       &                                                                                                                                                                       Remove all blue squares to win. &                 &               X &                   &                   &           &       + \\ \cline{2-8}
       &                                                                                                                                                                        You cannot move purple blocks. &                 &                 &                   &                 X &           &         \\ \hline
   Push Boulders &                                                                                                                                                                    Be careful not to box yourself in. &               X &                 &                   &                   &           &       - \\ \cline{2-8}
    &                                                                                                                                              Don't touch red/pink blocks, can grab light blue blocks. &                 &               X &                   &                 X &           &         \\ \cline{2-8}
    &                                                                                                                                                                          Try to get the yellow block. &                 &               X &                   &                   &           &         \\ \cline{2-8}
    &                                                                                                                                                                           Avoid pink \& orange blocks. &                 &               X &                   &                   &           &         \\ \cline{2-8}
    &                                                                                                                                                                 Later levels it is easy to get stuck! &                 &                 &                 X &                   &           &       - \\ \cline{2-8}
    &                                                                                                                                     You can use the green blocks to protect yourself by pushing them. &                 &                 &                 X &                   &           &         \\ \cline{2-8}
   &                                                                                               You can also push the green tiles into the pink tiles, but they are just moved and remain on the board. &                 &                 &                   &                 X &           &         \\ \hline
   Preconditions &                                                                                     Solve the puzzle by first gathering enough white boxes to pass through the green boxes and get to the yellow box. &               X &                 &                   &                   &           &       + \\ \cline{2-8}
    &                                                                                                                                                                       Don't waste the white you gain. &               X &                 &                   &                   &           &         \\ \cline{2-8}
    &                                                                                                                                             DO NOT touch a GREEN box without collecting enough WHITE! &               X &                 &                   &                   &           &         \\ \cline{2-8}
    &                                                                                                        Navigate to the white block and then go to the yellow one after passing through the green one. &                 &               X &                   &                   &           &         \\ \cline{2-8}
    &                                                                                                                                                            One white cube cancels out one green cube. &                 &                 &                   &                 X &           &         \\ \hline
    Plaqueattack &                                                                                                                                                                   Press the spacebar to fire bullets. &                 &               X &                   &                 X &           &         \\ \cline{2-8}
     &                                                           Defending your position from a far side of the map and shooting directly across will prevent enemy squares from reaching their destination. &                 &                 &                 X &                   &           &         \\ \cline{2-8}
     &                                                                                                                                                            The moving orange squares turn them green. &                 &                 &                   &                 X &           &         \\ \cline{2-8}
     &                                                                                                                                                          However I am not able to succeed in the game &                 &                 &                   &                   &         X &       - \\ \cline{2-8}
     &                                                                                                                                                                                               No idea &                 &                 &                   &                   &         X &         \\ \hline
        Lemmings &                                                                                                            Once YELLOW square is triggered it will seek out the nearest wall so don't be in its path. &               X &                 &                 X &                   &           &         \\ \cline{2-8}
         &                                                       its worth it at times to destroy some walls by punching holes to get a perfect system going making red squares easily go into the green square. &               X &                 &                 X &                   &           &         \\ \cline{2-8}
         &                                                                                                                               Let the red squares reach the green square in as few moves as possible. &               X &                 &                   &                   &           &         \\ \cline{2-8}
         &                                                                                                                                                                       You play as the dark blue tile. &                 &                 &                   &                 X &           &         \\ \cline{2-8}
         &                                                                                                                                                                         I don't understand how to win &                 &                 &                   &                   &         X &       + \\ \cline{2-8}
         &                                                                                                                                                                                    That's all I know. &                 &                 &                   &                   &         X &         \\ \cline{2-8}
         &                                                                                                                                           I was able to merge with the red but it did not collect it. &                 &                 &                   &                   &         X &         \\ \cline{2-8}
         &                                                                                                                                                               Not sure what green block is there for. &                 &                 &                   &                   &         X &         \\ \hline
 Explore/Exploit &                                                                                                                                                                   Collect all of the LEFT-MOST COLOR. &               X &                 &                   &                   &           &         \\ \cline{2-8}
  &                                                                                                                Move the DARK BLUE box to collect other colored boxes, based on the following pattern. &                 &               X &                   &                 X &           &         \\ \cline{2-8}
  &                                                                                                                                             If there are no red blocks, get all the blocks there are. &                 &               X &                   &                   &           &         \\ \cline{2-8}
  &                                                                                                         I.e, if there are two green squares and five yellow squares, eat the two green squares first. &                 &               X &                   &                   &           &         \\ \cline{2-8}
 &                                                                                                                                                                             Red, Orange, pink, green. &                 &                 &                   &                   &         X &         \\ \cline{2-8}
  &                                                                                                                                                                                            Good luck! &                 &                 &                   &                   &         X &         \\ \hline
    Avoid George
     &                                                                                                 Use the Space Bar to "shoot" a BROWN box from the BLUE box, in the last direction the BLUE box moved. &                 &               X &                   &                 X &           &         \\ \cline{2-8}
     &                                                                                                                                                               Do NOT let the yellow touch you either. &                 &               X &                   &                   &           &         \\ \cline{2-8}
     &                                                                                                                                                       Beware, the yellow square can go through walls. &                 &                 &                   &                 X &           &         \\ \cline{2-8}
     &                                                                                                                                                                                    Good luck comrade. &                 &                 &                   &                   &         X &         \\ \cline{2-8}
    &                                                                                                                                                                                I didn't try using it. &                 &                 &                   &                   &         X &         \\ 
\bottomrule
\end{tabular}
\small{\caption{A random sampling of utterances/sentences produced by participants for each game and their classification according to our language coding scheme.}} 

\end{center}
\end{table*}